\documentclass[runningheads]{llncs}
\usepackage{amsmath}
\usepackage{amssymb}
\usepackage{hyperref}
\usepackage[T1]{fontenc}
\usepackage{graphicx}
\usepackage{subcaption}
\usepackage{booktabs}
\usepackage{caption}
\begin{document}
\title{ZC-Swish: Stabilizing Deep BN-Free Networks for Edge and Micro-Batch Applications}
\author{Suvinava Basak\inst{}\orcidID{0009-0005-4289-1205}}
\institute{TU Braunschweig, Universitätspl. 2, 38106 Braunschweig, Germany
\email{suvinava.basak@tu-braunschweig.de}}
\maketitle
\begin{abstract}
Batch Normalization (BN) is a cornerstone of deep learning, yet it fundamentally breaks down in micro-batch regimes (e.g., 3D medical imaging) and non-IID Federated Learning. Removing BN from deep architectures, however, often leads to catastrophic training failures such as vanishing gradients and dying channels. We identify that standard activation functions, like Swish and ReLU, exacerbate this instability in BN-free networks due to their non-zero-centered nature, which causes compounding activation mean-shifts as network depth increases. In this technical communication, we propose \textbf{Zero-Centered Swish (ZC-Swish)}, a drop-in activation function parameterized to dynamically anchor activation means near zero. Through targeted stress-testing on BN-free convolutional networks at depths 8, 16, and 32, we demonstrate that while standard Swish collapses to near-random performance at depth 16 and beyond, ZC-Swish maintains stable layer-wise activation dynamics and achieves the highest test accuracy at depth 16 (51.5\%) with seed 42. ZC-Swish thus provides a robust, parameter-efficient solution for stabilizing deep networks in memory-constrained and privacy-preserving applications where traditional normalization is unviable.\footnote{Under review at KI 2026.}

\end{abstract}
\section{Introduction}
The optimization of deep neural networks heavily relies on Batch Normalization (BN)~\cite{ref1} to stabilize the distribution of internal activations. Coupled with modern smooth activation functions like Swish (SiLU)~\cite{ref2} or Mish~\cite{ref3}, BN enables training of increasingly deep architectures by mitigating covariate shift and ensuring healthy gradient flow. However, batch statistics introduce severe limitations in practical engineering paradigms. In high-resolution 3D medical imaging or video processing, limited GPU memory forces the use of micro-batches (sizes of 1 or 2), where BN mathematically breaks down or introduces destructive noise~\cite{ref4}. In Federated Learning, averaging BN statistics across edge devices with highly diverse, non-IID data corrupts the global model and presents a tangible privacy risk~\cite{ref5}.

When BN is removed, deep networks become extremely difficult to train. Standard activation functions like ReLU and Swish actively contribute to this instability: because they are not zero-centered, propagating data through them inherently shifts the activation mean in the positive direction. In deep, BN-free architectures, this positive mean-shift compounds layer by layer, ultimately leading to dead channels, vanishing gradients, and collapsed training. Workarounds such as learning rate warm-ups or Kaiming initialization~\cite{ref6} mask these issues rather than fixing them architecturally.

To address this at its root, we propose \emph{Zero-Centered Swish (ZC-Swish)}, a parameterized variant of Swish with learnable parameters that dynamically center the activation output. Through stress tests on BN-free PlainNets at depths 8, 16, and 32, we show that while standard activations collapse progressively with depth, ZC-Swish sustains meaningful learning up to depth 16 — positioning it as an effective, drop-in solution for memory-constrained and privacy-first domains.

\section{Related Work}
\textbf{Activation functions.} While ReLU~\cite{ref7} remains standard for simplicity, smoother alternatives like Swish~\cite{ref2} and Mish~\cite{ref3} offer superior generalization. However, all share a critical vulnerability: strictly non-negative or positively-biased outputs that shift activation means layer-by-layer, requiring normalization layers to continuously re-center the forward pass.

\textbf{Normalization alternatives.} Group Normalization~\cite{ref4} and Layer Normalization~\cite{ref8} compute statistics along channel or feature dimensions, bypassing BN's batch-size dependency. Though effective, they introduce inference overhead and can degrade accuracy. ZC-Swish internalizes the centering benefit directly in the activation, eliminating the need for separate spatial moment calculations.

\textbf{BN-free training strategies.} Fixup~\cite{ref9} and SkipInit~\cite{ref10} rescale residual branch initialization to prevent exploding gradients. NFNets~\cite{ref11} combine Adaptive Gradient Clipping with scaled weight standardization. However, these approaches depend on residual connections and complex initialization schemes. ZC-Swish is architecture-agnostic and imposes no such structural requirements.

\section{The Proposed Method: ZC-Swish}

\subsection{Mathematical Formulation}
Given input $x$, ZC-Swish is defined as:
\begin{equation}
f(x) = g \left[ (x-c)\,\sigma(\beta(x-c)) + c\,\sigma(-\beta c) \right]
\end{equation}
where $\sigma(\cdot)$ is the standard sigmoid, and $c$, $\beta$, $g$ are learnable parameters. The formulation guarantees \emph{origin preservation}: evaluating at $x=0$ shows the bias term exactly cancels the shift,
\begin{equation}
f(0) = g \left[ (-c)\,\sigma(-\beta c) + c\,\sigma(-\beta c) \right] = 0,
\end{equation}
ensuring inactive neurons contribute no positive baseline bias. The gradient preserves the smooth Swish landscape:
\begin{equation}
f'(x) = g \cdot \sigma(\beta(x-c)) \left[ 1 + \beta(x-c)(1 - \sigma(\beta(x-c))) \right].
\end{equation}

\begin{figure}[htbp]
    \centering
    \begin{subfigure}[b]{0.47\textwidth}
        \centering
        \includegraphics[width=\textwidth]{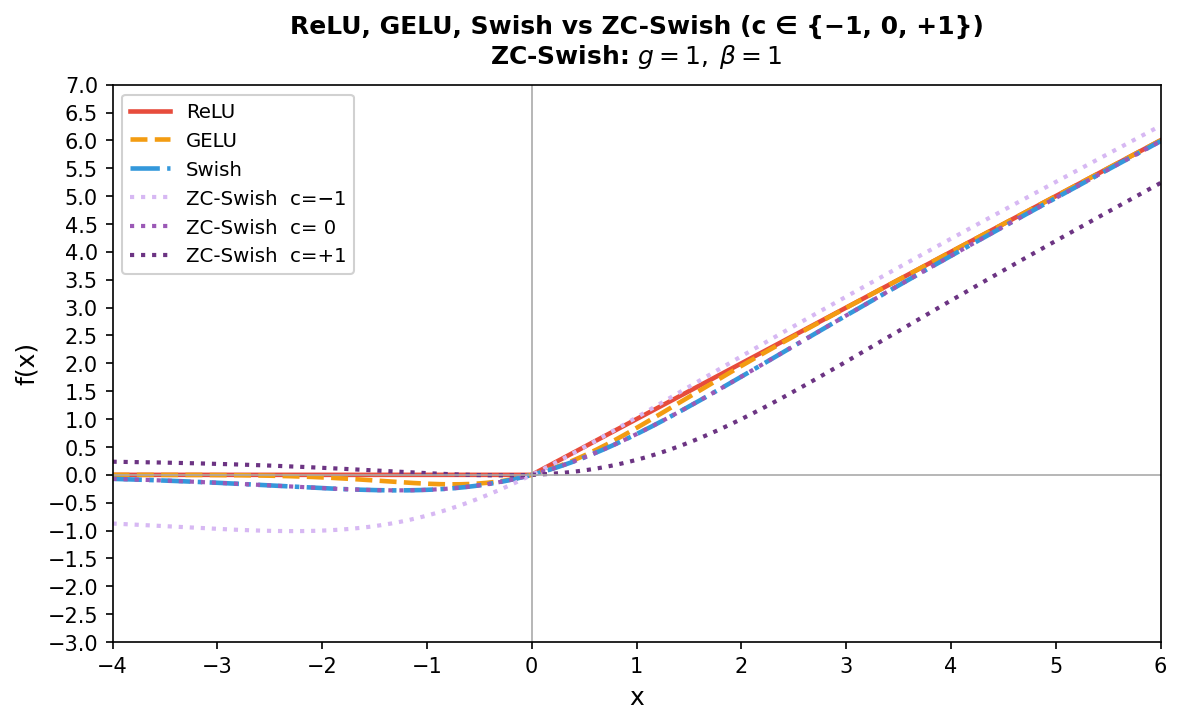}
        \caption{Comparison to ReLU, GELU, Swish}
        \label{fig:sub1}
    \end{subfigure}
    \hfill
    \begin{subfigure}[b]{0.47\textwidth}
        \centering
        \includegraphics[width=\textwidth]{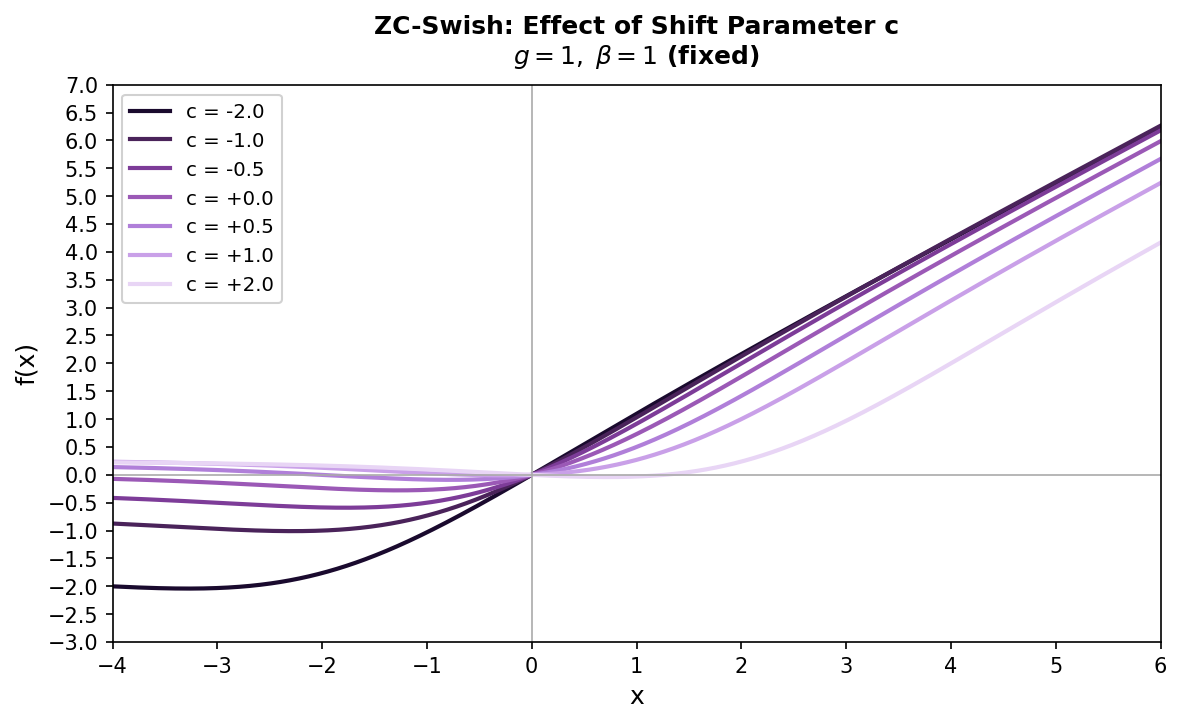}
        \caption{Effect of shift $c$}
        \label{fig:sub2}
    \end{subfigure}

    \vskip\baselineskip

    \begin{subfigure}[b]{0.47\textwidth}
        \centering
        \includegraphics[width=\textwidth]{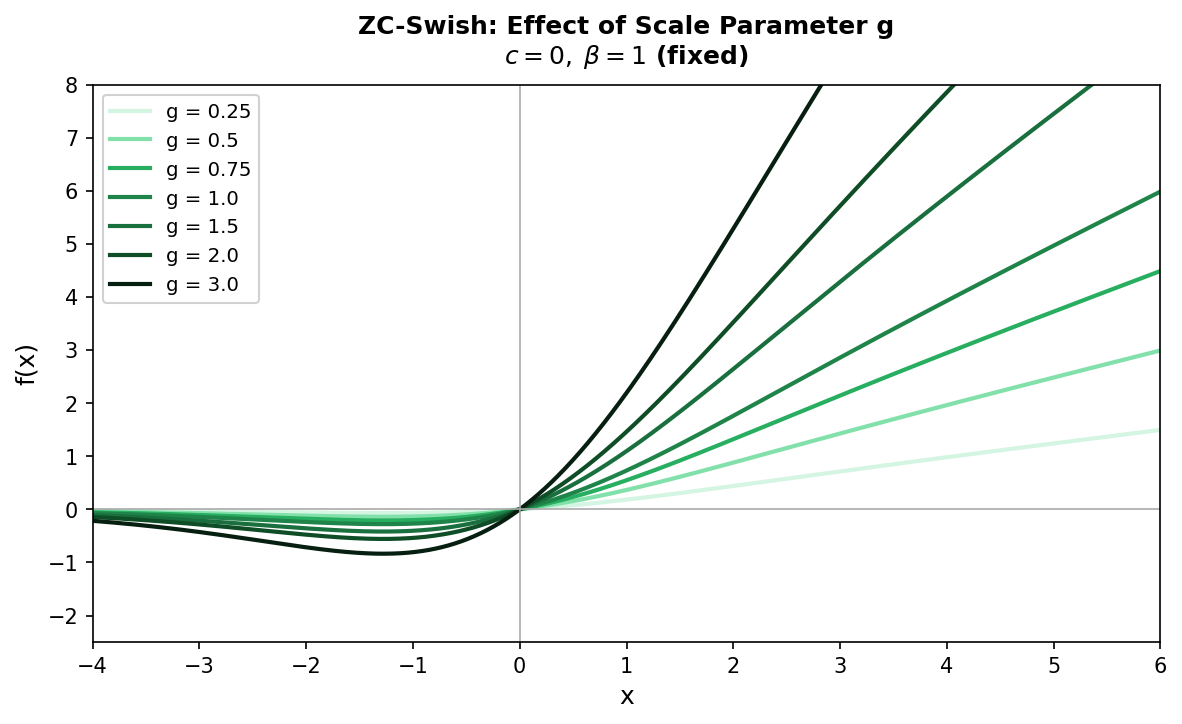}
        \caption{Effect of scale $g$}
        \label{fig:sub3}
    \end{subfigure}
    \hfill
    \begin{subfigure}[b]{0.47\textwidth}
        \centering
        \includegraphics[width=\textwidth]{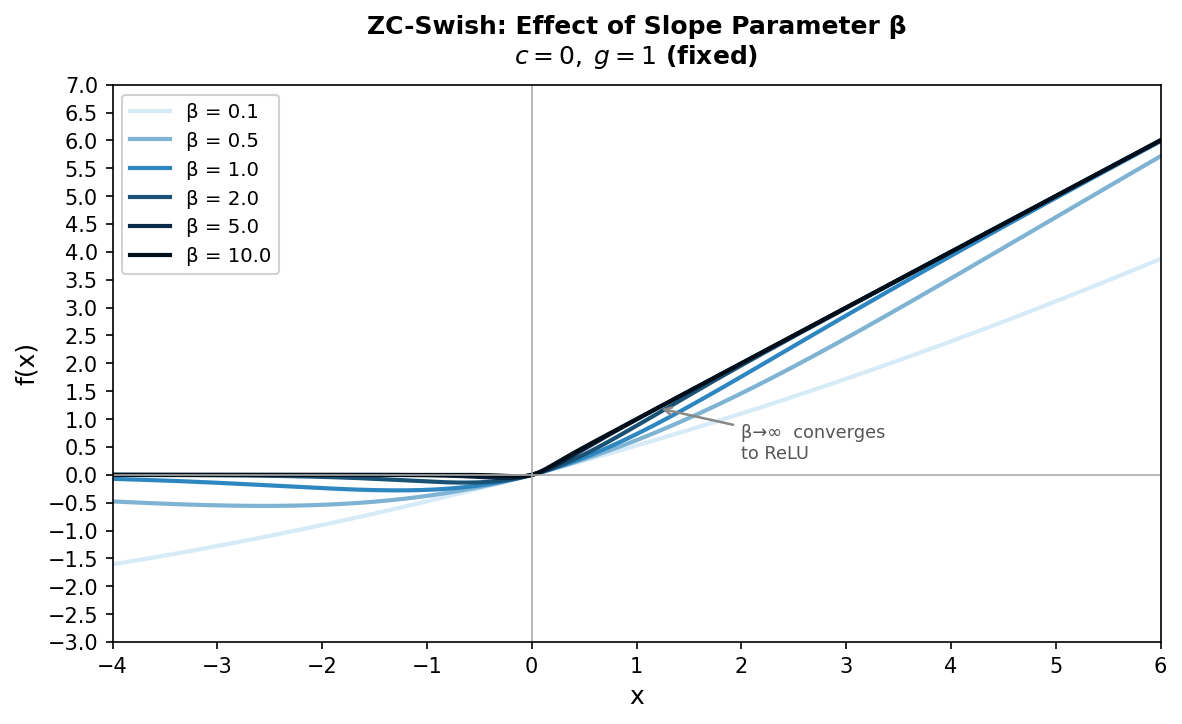}
        \caption{Effect of slope $\beta$}
        \label{fig:sub4}
    \end{subfigure}

    \caption{Comparison to baseline activations and the effect of $c$, $g$, $\beta$ on its shape.}
    \label{fig:baseline}
\end{figure}

Figure~\ref{fig:baseline} illustrates ZC-Swish against baseline activations and the effect of each learnable parameter.

\subsection{Learnable Parameters and Drift Mitigation}
ZC-Swish introduces three learnable parameters per channel: the \textbf{centering anchor} $c$ (initialized at $0.01$), which shifts the input space to counteract positive mean drift; the \textbf{steepness multiplier} $\beta$ (via $\beta = \text{softplus}(\beta_{raw})$, $\beta_{raw}=0.5413$, so $\beta \approx 1.0$ initially), ensuring $\beta>0$; and the \textbf{scale/gain} $g$ (initialized to $1.0$), which preserves activation variance at depth.

In standard BN-free Swish, the expected output is strictly positive ($\mathbb{E}[f(x)] > 0$), compounding exponentially across layers. ZC-Swish breaks this cycle: if activations drift positive, the network adjusts $c$ to restore $\mathbb{E}[f(x)] \approx 0$, internalizing the centering role of a normalization layer. For a PlainNet-16 with 15,028,644 parameters, ZC-Swish adds only 12,672 activation parameters ($0.084\%$ overhead), making it highly practical for edge deployments.

\section{Experimental Setup}
All experiments are conducted on \emph{CIFAR-100} ($100$ classes, $50000$ train / 10,000 test, $32{\times}32$) using a custom \texttt{PlainNet} — a VGG-style backbone with \emph{all} Batch Normalization and residual connections removed, maximizing sensitivity to activation mean-shift. The depth-16 config uses channel progression [64, 64, 128, 128, 256, 256, 256, 512, 512, 512, 512, 512, 512] with five MaxPool layers, plus a fully-connected head (Linear($512,512$) $\rightarrow$ ReLU $\rightarrow$ Dropout $\rightarrow$ Linear($512,100$)). We evaluate depths $D \in \{8, 16, 32\}$.

To isolate the activation function's native robustness, we deliberately withhold standard training stabilizers: PyTorch default uniform initialization (not Kaiming), AdamW (lr $=1{\times}10^{-3}$, weight decay $=5{\times}10^{-4}$), no learning rate warm-up, no cosine annealing, and 30 epochs per run. Batch size is 128. Seeds $\{42, 0, 12345\}$ are used for the depth-16 primary experiment; seed $\{42\}$ for depth scaling. All experiments ran on Google Colab with a Tesla T4 GPU (free tier). For depth-16, we report mean $\pm$ std of best-epoch accuracy across seeds; for depth scaling, best test accuracy per configuration.

\section{Results and Discussion}
\label{sec:results}

\subsection{Depth-16 Primary Results}
\label{subsec:depth16}

Table~\ref{tab:depth16} summarises results at depth 16 across three seeds. ZC-Swish achieves the highest mean training accuracy ($61.47 \pm 39.16\%$). On the test set, Swish attains a slightly higher mean ($43.73 \pm 9.25\%$) than ZC-Swish ($37.19 \pm 21.27\%$), but with lower variance, indicating ZC-Swish converges strongly on some seeds while remaining unstable on others. ReLU and GELU largely fail, achieving only $10.75\%$ and $17.63\%$ test accuracy respectively, near CIFAR-100's 1\% chance baseline.

\begin{table}[h]
  \centering
  \caption{Depth-16 performance --- mean $\pm$ std across 3 seeds.}
  \label{tab:depth16}
  \small
  \begin{tabular}{lrrrr}
    \toprule
    \textbf{Activation}
      & \textbf{Total Params}
      & \textbf{Act.\ Params}
      & \textbf{Best Train Acc.\ (\%)}
      & \textbf{Best Test Acc.\ (\%)} \\
    \midrule
    ReLU     & 15{,}028{,}644 & 0      & $12.13 \pm 15.79$ & $10.75 \pm 13.79$ \\
    GELU     & 15{,}028{,}644 & 0      & $24.77 \pm 33.53$ & $17.63 \pm 23.52$ \\
    Swish    & 15{,}028{,}644 & 0      & $57.04 \pm 19.38$ & $43.73 \pm\ \ 9.25$ \\
    \textbf{ZC-Swish} & \textbf{15{,}041{,}316} & \textbf{12{,}672}
      & $\mathbf{61.47 \pm 39.16}$ & $\mathbf{37.19 \pm 21.27}$ \\
    \bottomrule
  \end{tabular}
\end{table}

\begin{figure}[h]
  \centering
  \includegraphics[width=0.8\linewidth]{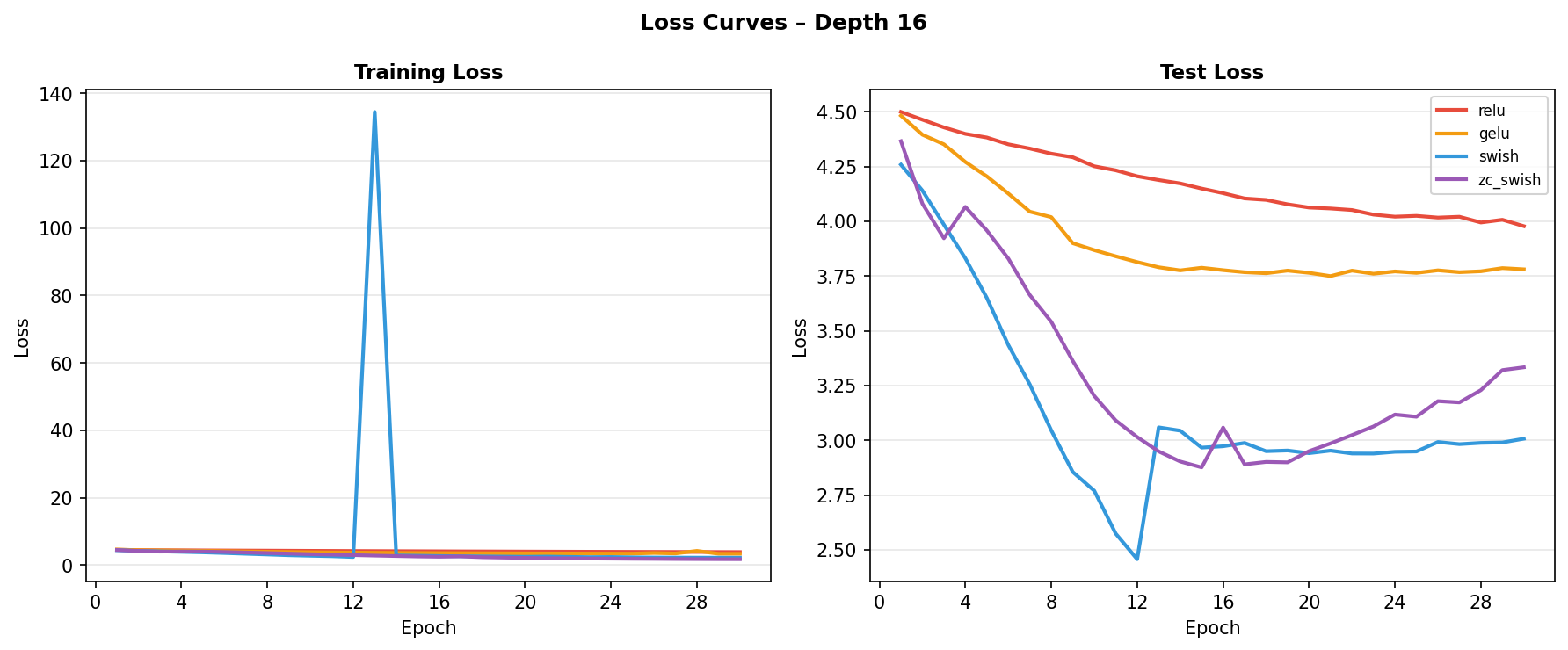}
  \caption{Training (left) and test (right) loss curves at depth~16.}
  \label{fig:loss_depth16}
\end{figure}

Figure~\ref{fig:loss_depth16} presents training and test loss curves at depth 16. Swish exhibits a sharp gradient explosion spike near epoch 13 (training loss $>130$), likely caused by accumulated positive activation drift, though it partially recovers to a test loss near 3.0. ZC-Swish avoids such spikes and achieves lower test loss by mid-training, consistent with its zero-centering design. ReLU and GELU show flat loss curves throughout, indicative of network collapse.

\subsection{Depth Scaling Results}
\label{subsec:depth_scaling}

\begin{figure}[h]
  \centering
  \includegraphics[width=\linewidth]{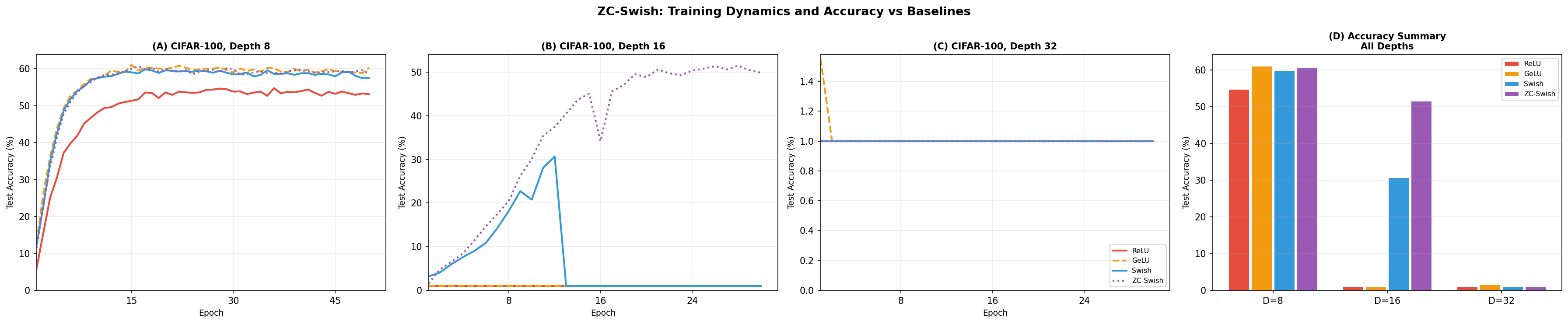}
  \caption{Training dynamics and accuracy summary across depths. (A)~Depth~8; (B)~Depth~16; (C)~Depth~32; (D)~Accuracy summary.}
  \label{fig:depth_combined}
\end{figure}

Table~\ref{tab:depth_scaling} and Figure~\ref{fig:depth_combined} present depth-scaling results (single seed, 30 epochs) across $D \in \{8, 16, 32\}$.

\begin{table}[h]
  \centering
  \caption{Best test accuracy (\%) by activation and depth (single seed).}
  \label{tab:depth_scaling}
  \small
  \begin{tabular}{lccc}
    \toprule
    \textbf{Activation} & \textbf{Depth 8} & \textbf{Depth 16} & \textbf{Depth 32} \\
    \midrule
    ReLU              & 54.7 & \phantom{0}1.0 & 1.0 \\
    GELU              & 61.0 & \phantom{0}1.0 & 1.6 \\
    Swish             & 59.9 &            30.7 & 1.0 \\
    \textbf{ZC-Swish} & \textbf{60.7} & \textbf{51.5} & \textbf{1.0} \\
    \bottomrule
  \end{tabular}
\end{table}

At \textbf{depth 8} (Panel A), all activations converge comparably (ReLU: 54.7\%, GELU: 61.0\%, Swish: 59.9\%, ZC-Swish: 60.7\%), confirming activation drift is negligible at shallow depths.
At \textbf{depth 16} (Panel B), ReLU and GELU collapse to 1.0\% while Swish reaches 30.7\% despite an instability spike. ZC-Swish significantly outperforms all baselines at \textbf{51.5\%}, a gap of $+20.8$ percentage points over Swish, demonstrating effective mean-shift mitigation.
At \textbf{depth 32} (Panel C), all activations including ZC-Swish collapse to ${\approx}1.0\%$, indicating that 30 epochs with default initialization and no residual connections exceeds the current stabilizing capacity of ZC-Swish. Panel D summarizes this trend clearly: comparable at depth 8, strong differentiation at depth 16 in favour of ZC-Swish, universal collapse at depth 32. Further investigation with extended training budgets and improved initialization is warranted.

\section{Conclusion}
We proposed ZC-Swish, a lightweight parameterized activation function that addresses compounding mean-shift in deep BN-free networks via three learnable parameters ($c$, $\beta$, $g$) that preserve $f(0)=0$ and dynamically anchor activations near zero, adding only $0.084\%$ parameter overhead. Stress tests on CIFAR-100 BN-free PlainNets confirm three findings: (1)~at depth 8, all activations perform comparably; (2)~at depth 16, ZC-Swish achieves 51.5\% test accuracy (single seed) and 61.47\% mean training accuracy, substantially outperforming Swish (30.7\%), ReLU (1.0\%), and GELU (1.0\%); (3)~at depth 32, all activations collapse under the current training regime. ZC-Swish is a practical drop-in solution for BN-free settings such as micro-batch and Federated Learning deployments. Future work should explore depth 32 with extended training budgets, improved initialization strategies, and residual architectures to fully characterize its stabilization limits.

\section*{Limitations}
The depth-32 collapse may stem from insufficient training (30 epochs), suboptimal default initialization, or the absence of residual connections, factors requiring further ablation. These constraints arise from conducting all experiments independently on Google Colab's free-tier Tesla T4 GPU, which limited the number of epochs, seeds (3 for primary experiment with depth 16), and hyperparameter searches. Additionally, evaluation is restricted to CIFAR-100; broader validation on ImageNet, medical imaging datasets, and transformer-based BN-free architectures remains future work. The current results should therefore be interpreted as a proof-of-concept in a controlled stress-test environment, rather than a comprehensive evaluation across all scenarios. A more comprehensive evaluation with extended compute would reduce the variance in the reported results and allow systematic tuning of ZC-Swish's initialization strategy at higher depths.

\begin{credits}

\subsubsection{Reproducibility and availability.} The full system code is publicly available on \href{https://github.com/baksho/zc-swish}{GitHub}. The repository is released under the CC BY-NC-SA 4.0 license.
\subsubsection{\ackname} The author declares that no external funding was received for this work. This research received no specific grant from any funding agency in the public, commercial, or not-for-profit sectors.

\subsubsection{\discintname}
The author have no competing interests to declare that are relevant to the content of this article.
\end{credits}
%
%

\end{document}